# Hidden Trends in 90 Years of Harvard Business Review

Chia-Chi Tsai Chao-Lin Liu Wei-Jie Huang Man-Kwan Shan
Department of Computer Science
National Chengchi University, Taipei, Taiwan
{99753006, chaolin, 100753014, mkshan}@nccu.edu.tw

## Abstract
In this paper, we demonstrate and discuss results of our mining the abstracts of the publications in *Harvard Business Review* between 1922 and 2012. Techniques for computing n-grams, collocations, basic sentiment analysis, and named-entity recognition were employed to uncover trends hidden in the abstracts. We present findings about international relationships, sentiment in HBR's abstracts, important international companies, influential technological inventions, renown researchers in management theories, US presidents via chronological analyses.

## 1. Introduction
Digital humanities is an interdisciplinary area where researchers employ computing technologies for the research and applications of subjects in humanities. Relatively, digital humanities is a burgeoning area, but thrives rapidly in recent years [1][5].

Digital history [2] is a very important field in digital humanities. Seto et al. applied image processing techniques to rebuild the landscape of the Kyoto city and studied the development of the city [12]. Lee and Lee analyzed 300 hundred years of Korean culinary manuscripts [8]. Some researchers of Chinese history have studied important and interesting historical events with the help of techniques of natural language processing (NLP), e.g., [10], in the past few years. Jin et al. study the conceptualization of "Chinese people" ("華人", /hua2 ren2/) in the modern Chinese history [8].

In this paper, we showcase applications of the NLP techniques to the analysis of a well-known international magazine. We analyzed the abstracts of the articles published in the *Harvard Business Review* [6] (HBR, henceforth) between 1922 and 2012. Without the help of computing tools, it would not be easy to analyze the contents of thousands of abstracts and figure out the trends hidden in the text materials. The abstracts were not public, but became available to us when we participated in an activities that were administered by the Kaggle company[1].

In the following presentation and charts, we illustrate some statements that were not explicitly offered in HBR. We discovered some hidden information by mining the contents of the abstracts, with the help of NLP tools and some extra data that were not part of the HBR abstracts. The extra data include a Harvard dictionary for sentimental words, which classify words into positive and negative words [7], data about the annual GDP growths of USA [3], and memberships of the G20 [4] and OECD [11] countries.

We list the goals of our exploration in Section 2, and explain the methods for processing the text materials in Section 3. Findings regarding the initial goals were presented in several subsections in Section 4 .

## 2. Goals
For all issues listed below, we show some interesting trends that were hidden in the HBR abstracts through charts of the annual changes of selected metrics.

We explored six interesting questions.

1. When USA was mentioned in HBR, what countries of other regions were mentioned at the same time? This provides indicators for how USA cooperated with other countries. We used computer tools to determine whether an article discussed issues related to USA and other countries. The counts for individual countries of a region, e.g., the Asia Pacific region, are added for the region.
2. Is HBR a "CALM" periodical? We employed a Harvard dictionary of sentimental words (a Harvard dictionary of sentimental words) to find "positive" and "negative" words in HBR. We studied the annual percentages of positive and negative words in HBR articles.
3. Which and when important elites were mentioned in the HBR abstracts? We studied some renown researchers in management theories and the US presidents.
4. HBR offers a window for watching the "important" companies for our lives. We analyzed the annual frequencies of selected companies that were mentioned in HBR abstracts.
5. HBR also provides a clue for when technical products come into our lives. By calculating the frequencies of some technological keywords in HBR, we looked back on how those products and technologies found their ways into our lives.
6. Of course, a good illustration needs a good graphical presentation.

## 3. Corpus and Methods
The original data file is a plain Excel files. The first row explains the types of data in individual columns. Each following row is an abstract, including the abstract and meta data about the abstract. The meta data include the publication date, the authors, the subject categories, etc.

For the current study, we extracted the bodies of the abstracts and saved each abstract into a file. Abstracts which were published in the same year were stored in the same folder, using the year as the name of the folder. Through this process, we obtained 13,522 abstracts which contained 1,536,564 words.

[1] http://www.kaggle.com

After saving the abstracts in folders for years, it was easy to compute the frequencies of annual frequencies of keywords and bigrams. Two keywords (a bigram) were considered to appear together if the same abstract contained them. Namely, the collocation window is the whole abstract. We then plot such chronological data in the illustrations. The analysis of annual frequencies of keywords is the same the Google Trend service[2]. The analysis of n-grams is similar to the Google Trend, but considers the co-occurrences of multiple keywords. For historical studies, the co-occurrences of keywords carry more precise semantics and the occurrences of individual keywords.

We employed the Stanford NLP tools[3] to extract named entities in the abstracts and to parse the sentences in the abstracts. The named entities recognition (often referred as NER) task helps to find names of countries and companies. The parsing of sentences would allow us to conduct deep understanding of the statements. However, we have not completed our investigation in this avenue.

We adopted the Harvard emotion dictionary to count the frequencies of positive and negative words. It is certainly possible to build a classifier to categorize the abstracts into positive and negative instances, but we have not done so at this moment.

# 4. The Hidden Trends

## 4.1 International Relationships

HBR mentioned USA with European and Asian countries more often during tough times. Figures 1 and Figure 2 indicate that, when USA was discussed, Asia Pacific and European countries were more frequently mentioned during World War II and during the financial crisis in recent years. Over other periods, HBR seldom mentioned USA and countries in multiple region in the same abstract during normal times. This phenomenon repeats when we study the G20 and the OCED countries.

Figures 3 and 4 show that Japan and Russia were special in that they were very important for USA's interests for a specific period. Their co-occurrences with USA in HBR betwen 1985 and 1995 were much more frequent than any other G20 and OECD countries and even regions.

## 4.2 HBR is a Calm Periodical

HBR abstracts use positive and negative words almost constantly over the years, irrespectively of the changes of annual GDP changes of USA. The curves in Figure 5 show the percentages of positive and negative words in a year's abstracts. The curve for the USA GDP shows the year over year change of GDP. The percentages of positive and negative words varied within a very narrow range, in contrast of the changes of USA GDP numbers.

Figure 6 shows the average numbers of positive and negative words in an HBR article. The curves show that the numbers of words are increasing slowly.

Notice that we did not classify the sentiment of individual abstracts. In that case, we must handle the negation words or double negations in statements. We simply counted the frequencies of positive and negative words at this moment.

## 4.3 Elites in HBR

We computed the annual frequencies of "Peter F. Drucker", "Peter M. Senge", "John P. Kotter", "Michael Porter", "Gary Hamel", "Henry Minzberg", "Chris Zook", "Tom Peters", "Noel Tichy", and "Gordon Donaldson" in the abstracts. When computing the frequencies, we consider a name with and without the middle names, e.g. we treated "Peter F. Drucker" and "Peter Drucker" as the same person.

One problem that we did not solve is that we did not distinguish two persons of the same name. If there was an ordinary person whose name is also "Peter Drucker" and if he was also mentioned in HBR, we would not obtain the correct frequencies for "Peter Drucker".

Peter Drucker was the leader in the group, and stated to be mentioned in HBR since 1950s. Figure 7 shows the annual frequencies of the five most frequently mentioned researchers since 1980. We can see that Peter Drucker is still the most popular researcher, but different researcher may become popular during specific periods. For instance, Gary Hamel was the most between 1989 and 1991, when the US economy was in depression.

It was interesting for us to check whether an abstract would mention two elites, which may imply that theories of two elites are discussed for an event. Unfortunately, the most frequently "co-mentioned" elites took place only four times for Peter Drucker and Peter Senge in 1998. Hence, we do not show a chart for this exploration.

We were interested in when HBR mentioned the US presidents and the chairpersons of the Federal Reserve. Figure 8 shows the frequencies of the US presidents in HBR since 1981. There are six presidents in this period. We did not use very distinguishable colors for different presidents in the chart because the main focus is when the presidents were mentioned in HBR. It is interesting for us to notice that it is the tough times when the US presidents were mentioned in HBR. For instance, in 1987, 1988, 2001, and recent years. There was a "Black Monday" in late 1987, the "burst of Internet bubble" around 2001, and global economic problems in recent years.

We did not show a chart for when HBR mentioned the chairpersons of the Federal Reserve. HBR rarely mentioned the chairpersons.

[2] http://www.google.com/trends/
[3] http://nlp.stanford.edu/software/

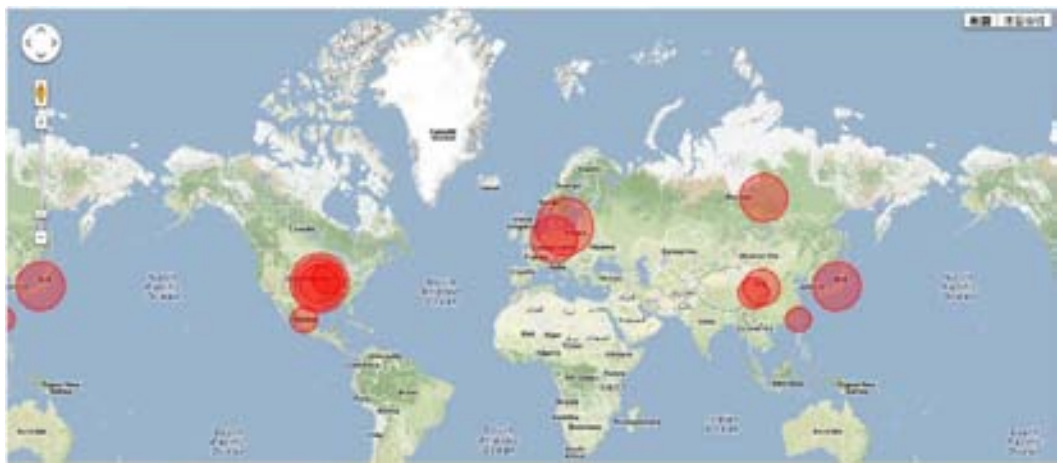

Figure 14. An interactive map

## 4.4 Well-Known Companies in HBR

Figures 9, 10, and 11 demonstrate the appearances of some big companies in HBR's abstracts. The vertical bars show the frequencies of the names of the companies in HBR. Apple, Google, and Samsung are in Figure 9. These three companies are “interacting” in multiple senses in recent years. HP, IBM, Intel, and Microsoft are in Figure 10. Ford, GE, GM, Bells (including Western Electric, AT&T, and Lucent) are in Figure 11. From Figure 10 and 11, we can see how the births and cycles of rises and falls of some well-known companies.

## 4.5 Technologies in our Lives

Again, the vertical axes of the charts in Figures 12 and 13 show the frequencies of the technological words in HBR. Computers appear to be the most lasting and influential products for our lives so far. Telephones and radios were very old technological products, as indicated in Figure 12.

Figure 13 may remind us the impact of “Internet”. The frequency of "Internet" dwarfs the frequencies of other popular words.

## 4.6 An Interactive Map

Finally, the visual impact would be greatly strengthened if we could show the charts with a map. The map in Figure 14 shows the locations of the top 10 most mentioned country names in HBR during 1990s. You can play with an interactive version at **http://cs.nccu.edu.tw/~chaolin/map.html**.

## 5. Concluding Remarks

We applied techniques for natural language processing (NLP) to the analysis of the contents of Harvard Business Review. The results show that NLP techniques are practically useful for business informatics. We believe that the charts uncovered some interesting trends in the HBR abstracts. However, a serious researcher does not stop at the point of finding these interesting trends. Experts would use the trends as seeds for furthering investigation about the past events. Hopefully some new discoveries can follow.

## Acknowledgements

We thank the valuable comments of Professor Ming-Feng Tsai on this study.

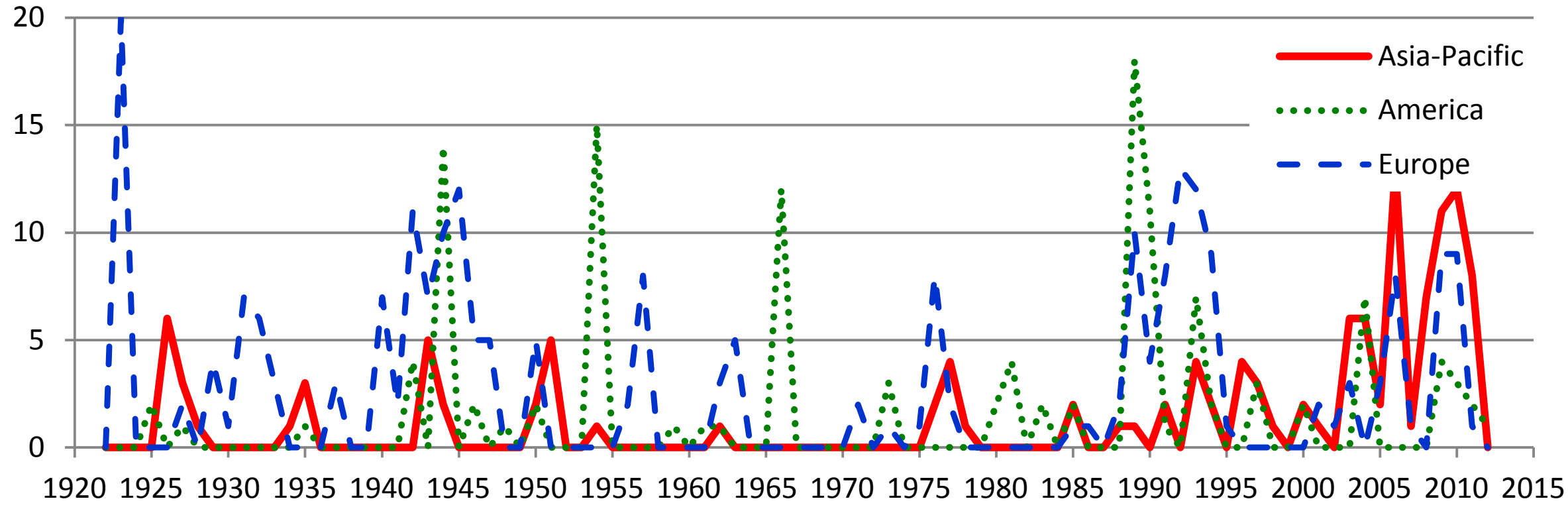

Figure 1. Annual frequency of USA being mentioned with G20 countries in three regions

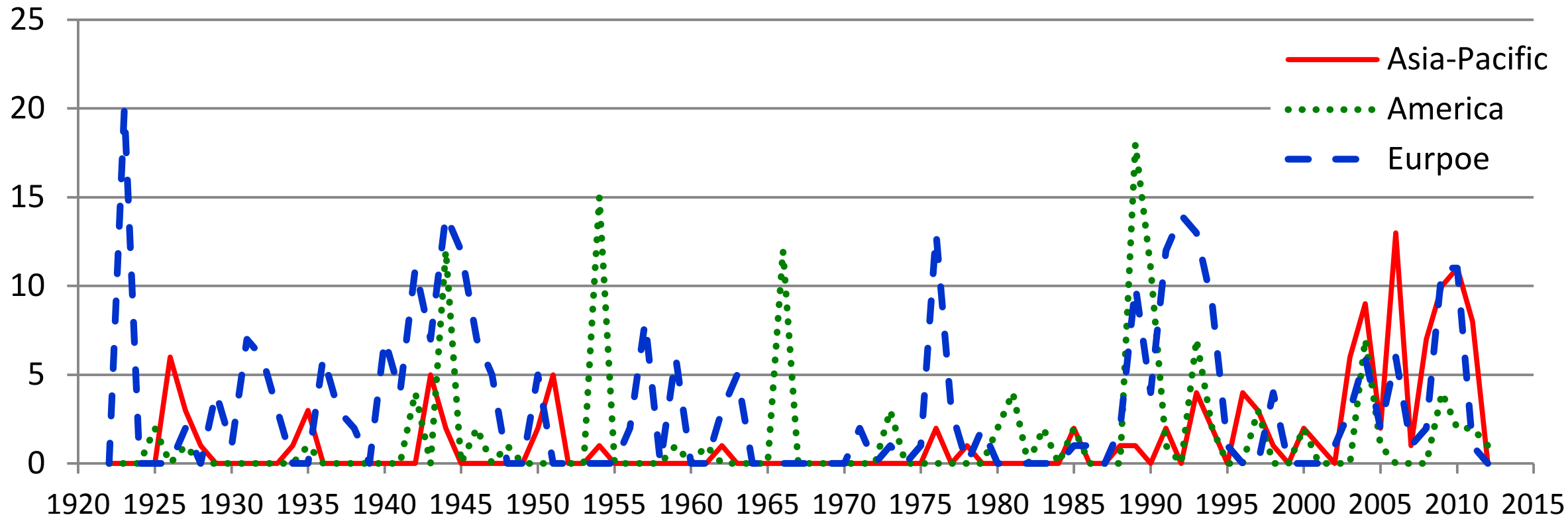

Figure 2. Annual frequency of USA being mentioned with OECD countries in three regions

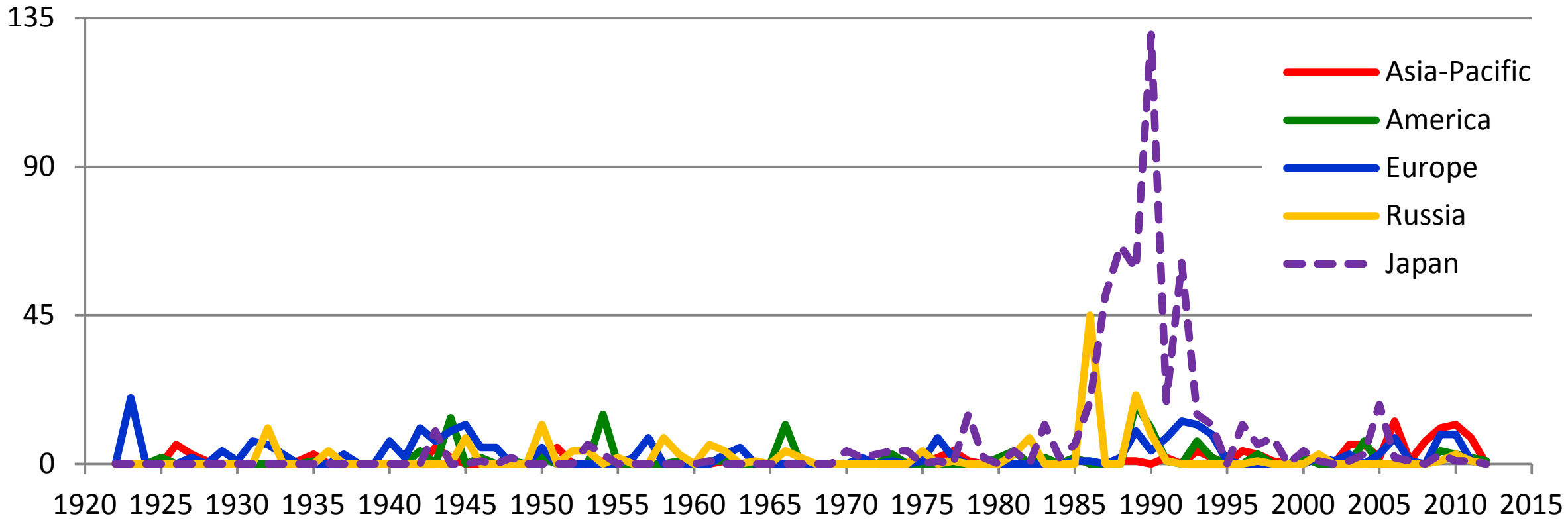

Figure 3. Annual frequency of USA being mentioned with Russia, Japan, and G20 countries

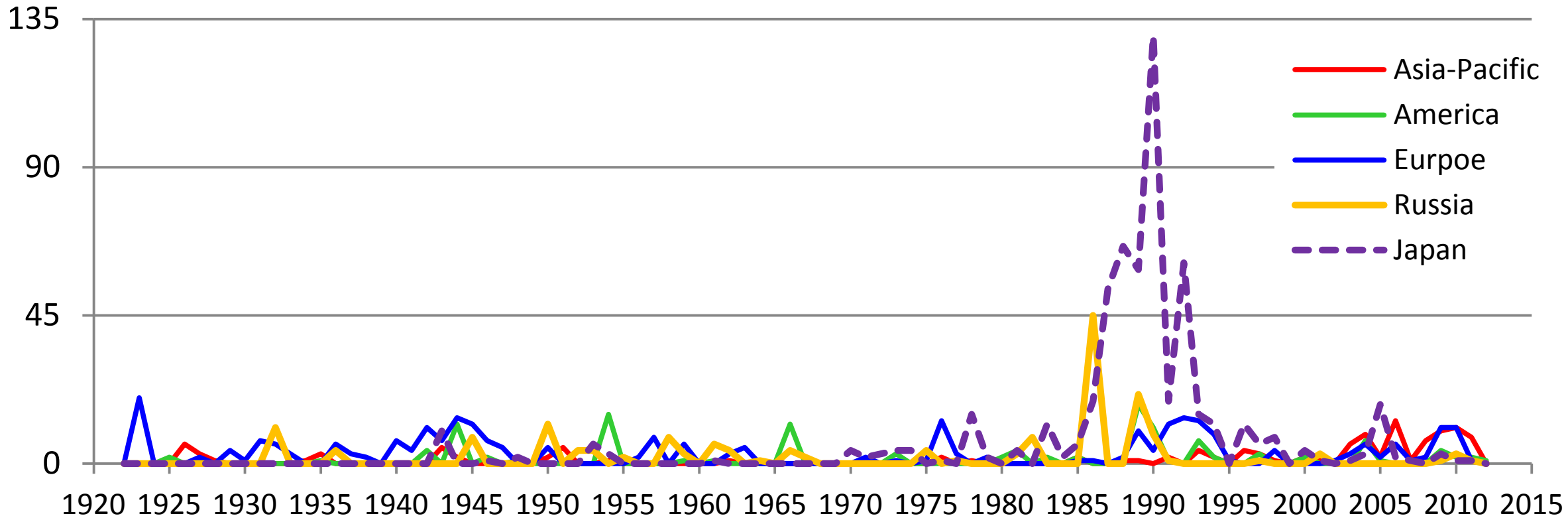

Figure 4. Annual frequency of USA being mentioned with Russia, Japan, and OECD countries

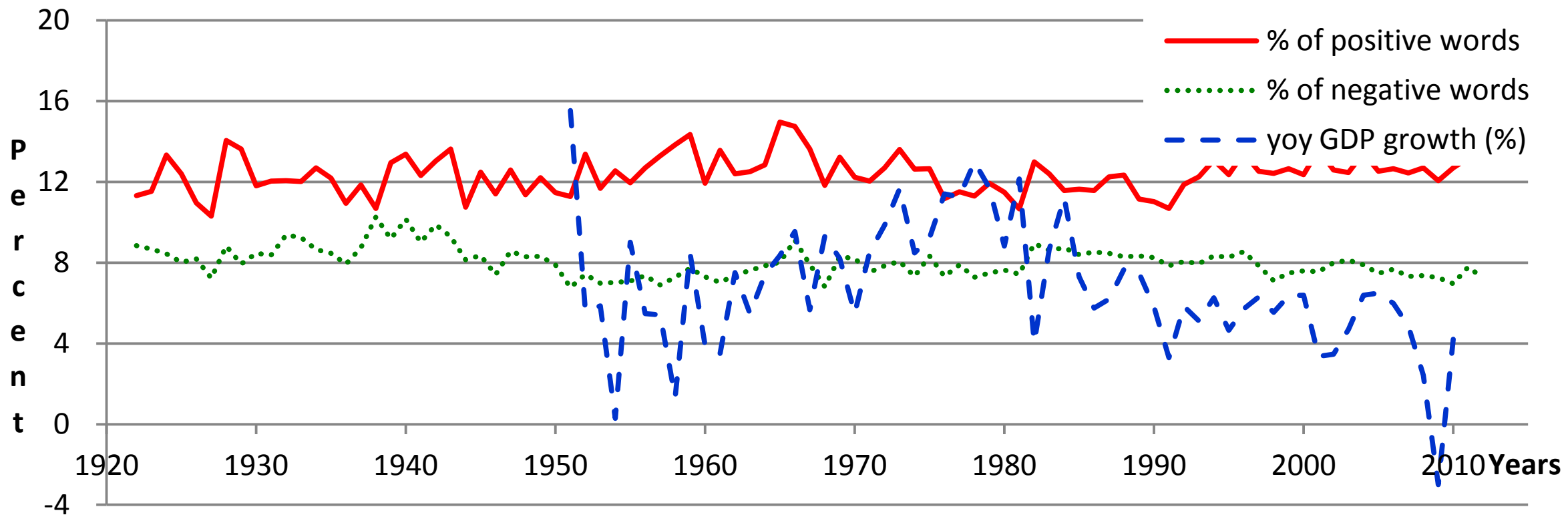

Figure 5. Sentiment of HBR remains relatively stable irrespective of USA GDP growth

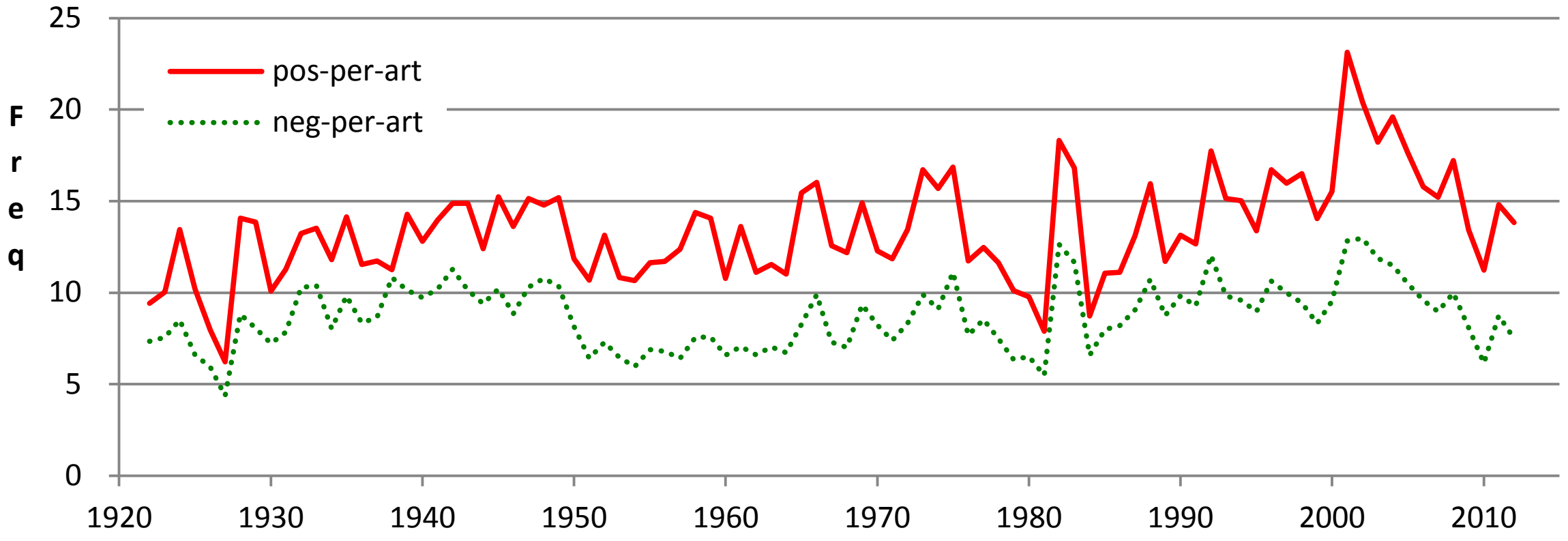

Figure 6. Numbers of positive/negative words per article changed gradually

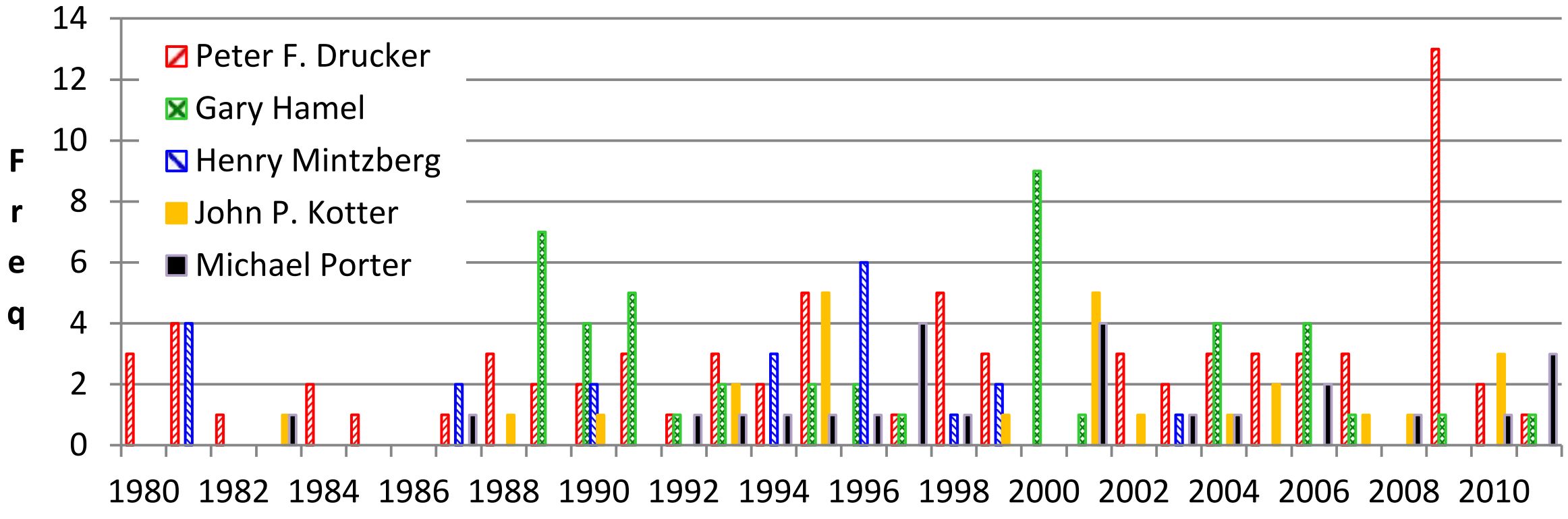

Figure 7. Famous researchers in management theories

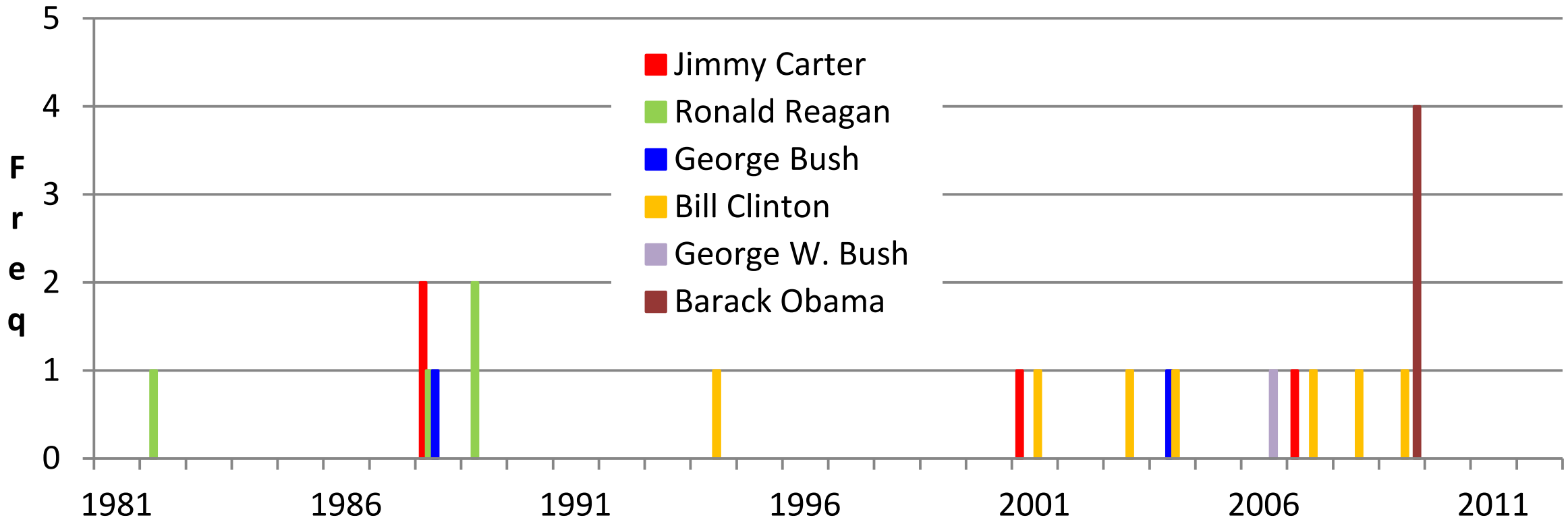

Figure 8. When HBR mentioned US Presidents

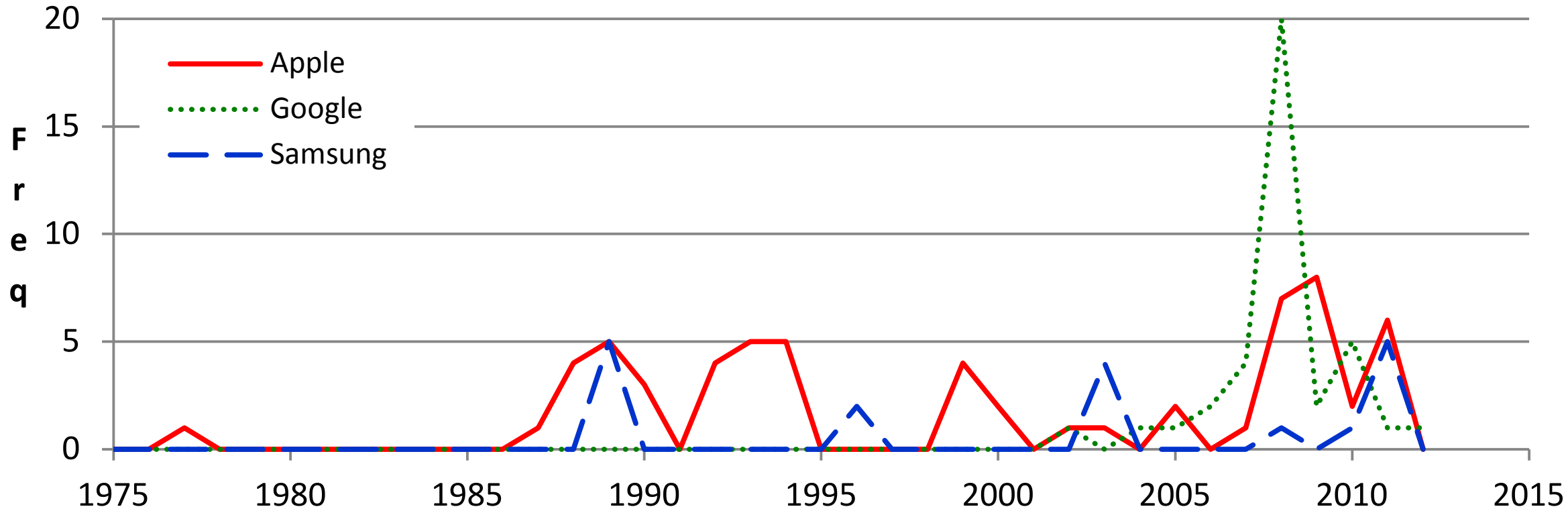

Figure 9. When some companies started to interact

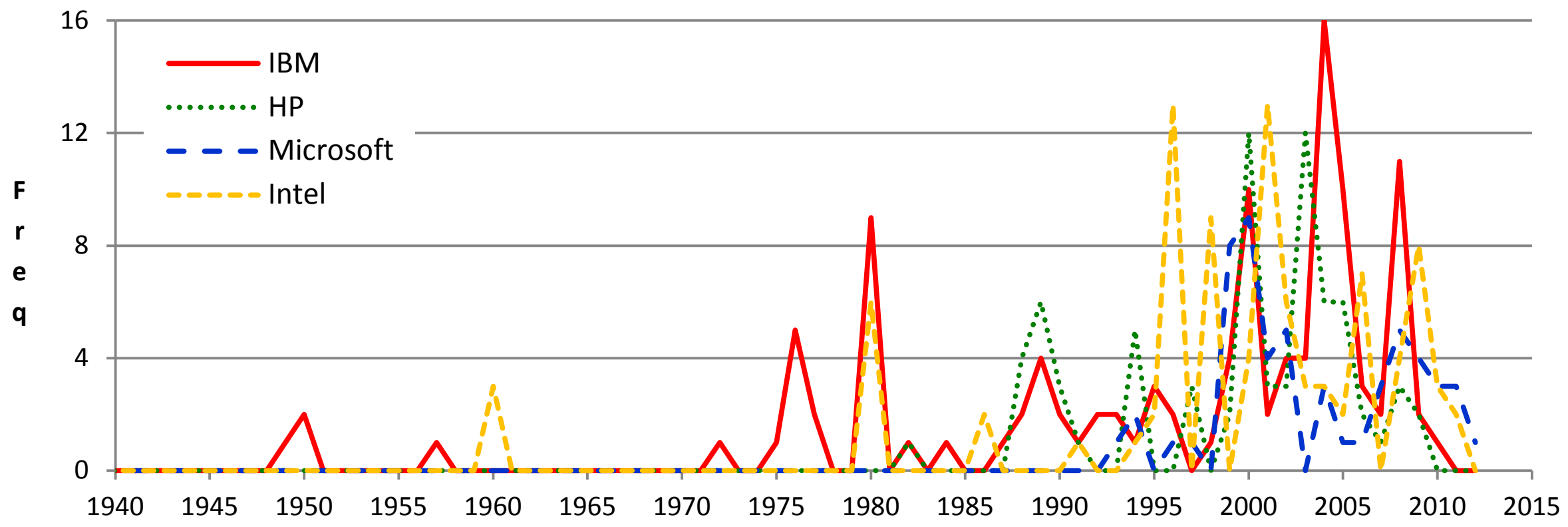

Figure 10. When some modern companies took the stage

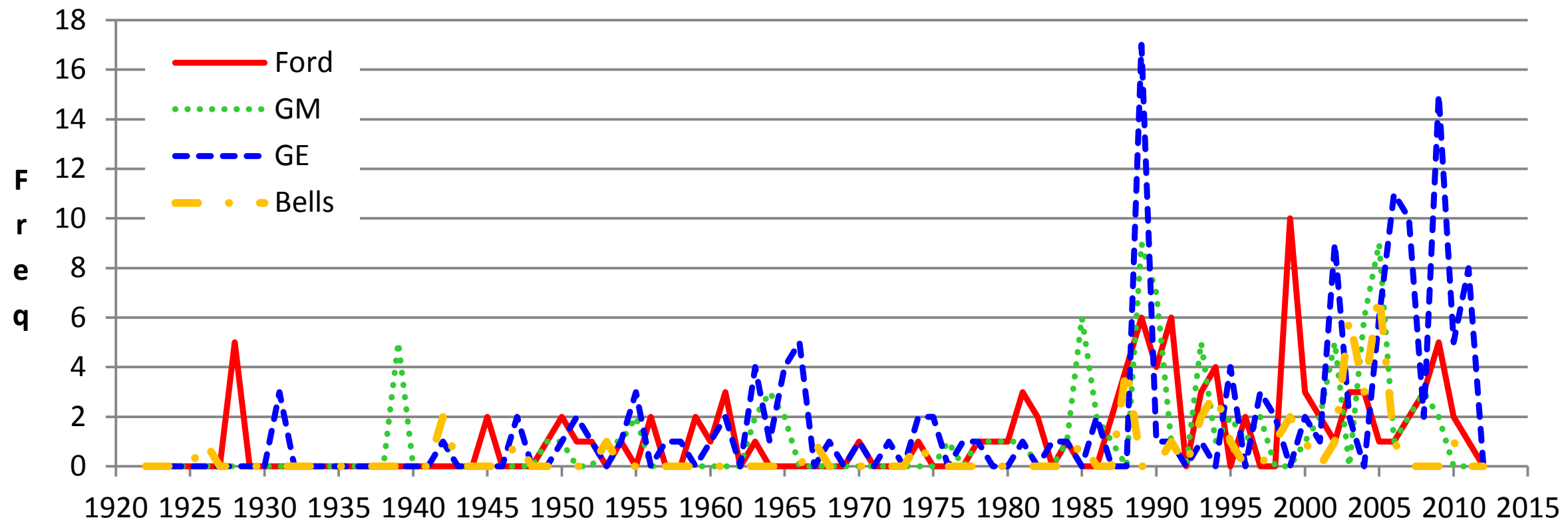

Figure 11. Cycles of some hundred-year old companies

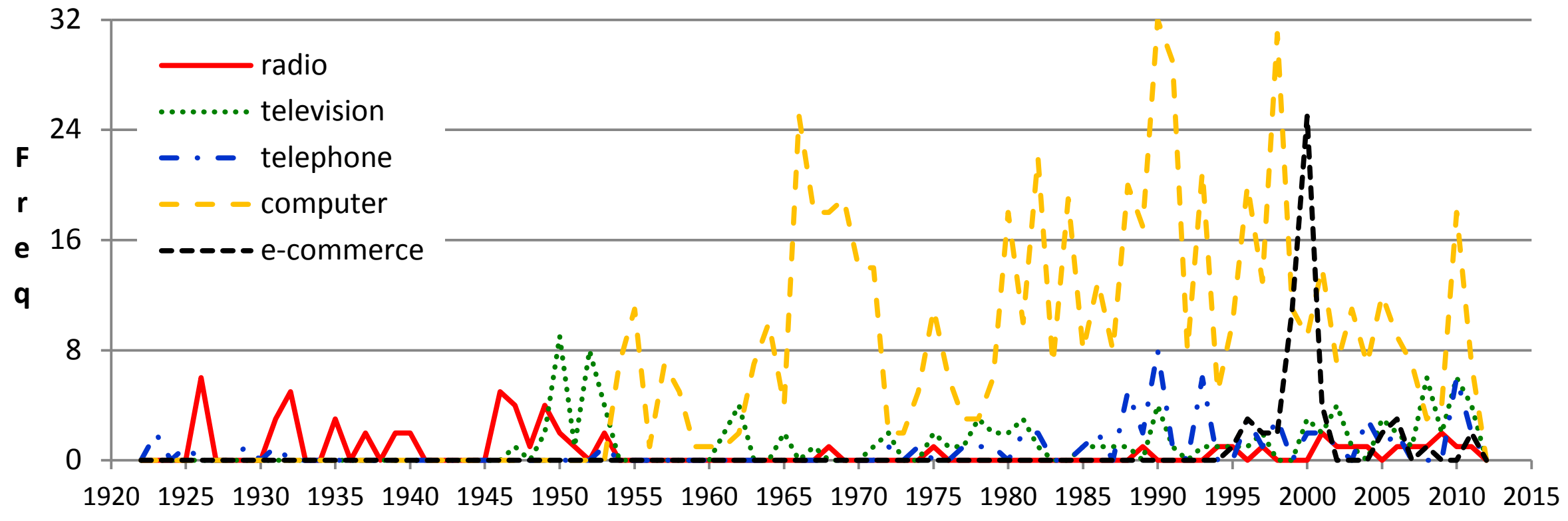

Figure 12. Some modern technologies in our lives

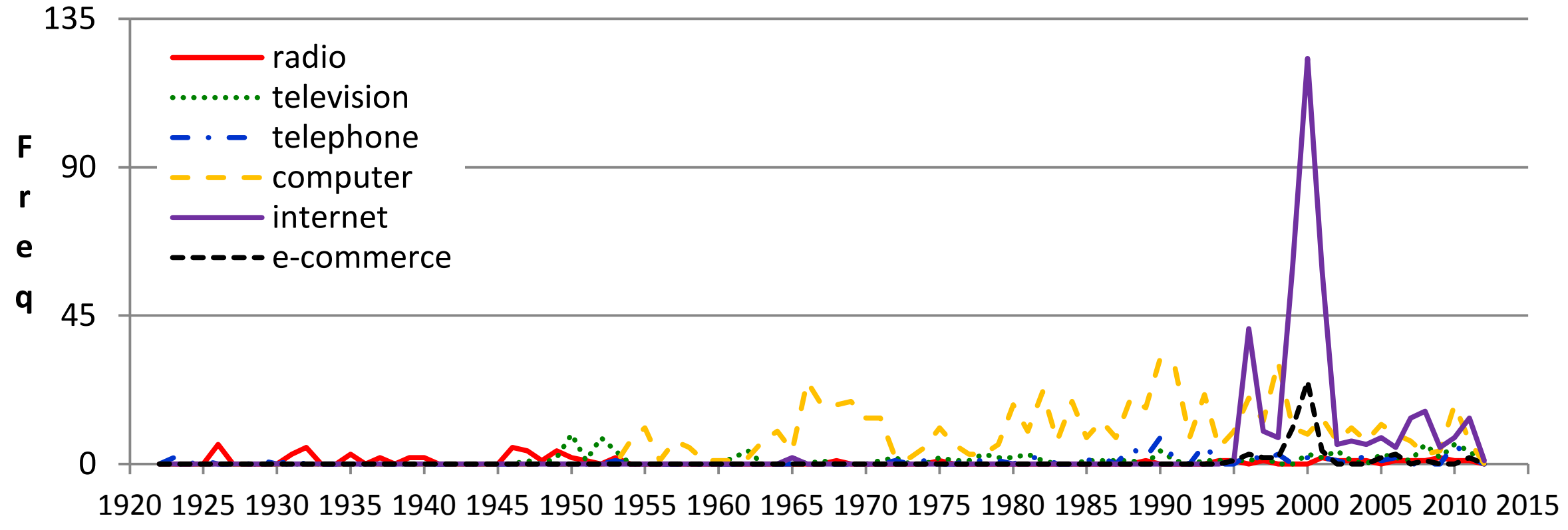

Figure 13. “internet” is dominating